\documentclass{article}
\usepackage{spconf,amsmath,graphicx}
\usepackage{algorithm}
\usepackage{algorithmic}
\usepackage{multirow}
\usepackage{arydshln }
\usepackage{amssymb}
\usepackage{amsmath}
\usepackage{booktabs}
\usepackage{enumitem}
\usepackage{xcolor} 
\usepackage{colortbl}
\usepackage[hidelinks]{hyperref}

\definecolor{shadecolor}{rgb}{0.92, 0.92, 0.92}
\definecolor{gtgray}{gray}{0.97}
\definecolor{mygray}{gray}{.88}

\definecolor{gray1}{gray}{.90}
\definecolor{gray2}{gray}{.92}
\definecolor{gray3}{gray}{.94}

\makeatletter
\def\hlinew#1{%
  \noalign{\ifnum0=`}\fi\hrule \@height #1 \futurelet
   \reserved@a\@xhline}
\makeatother

\title{Adaptive Super Resolution For One-Shot Talking-Head Generation}
\name{Luchuan Song$^{1}$ \qquad Pinxin Liu$^{1}$ \qquad Guojun Yin$^{2}$ \qquad Chenliang Xu$^{1}$
\address{$^1$University of Rochester \qquad $^2$University of Science and Technology of China}}

\begin{document}

\maketitle
\begin{abstract}
The one-shot talking-head generation learns to synthesize a talking-head video with one source portrait image under the driving of same or different identity video. Usually these methods require plane-based pixel transformations via Jacobin matrices or facial image warps for novel poses generation. The constraints of using a single image source and pixel displacements often compromise the clarity of the synthesized images. Some methods try to improve the quality of synthesized videos by introducing additional super-resolution modules, but this will undoubtedly increase computational consumption and destroy the original data distribution. In this work, we propose an adaptive high-quality talking-head video generation method, which synthesizes high-resolution video without additional pre-trained modules. Specifically, inspired by existing super-resolution methods, we down-sample the one-shot source image, and then adaptively reconstruct high-frequency details via an encoder-decoder module, resulting in enhanced video clarity. Our method consistently improves the quality of generated videos through a straightforward yet effective strategy, substantiated by quantitative and qualitative evaluations. The code and demo video are available on: \url{https://github.com/Songluchuan/AdaSR-TalkingHead/}.

\end{abstract}
\section{Introduction}
\vspace{-0.2cm}
\begin{figure*}[t]
    \centering
    \includegraphics[width=.9\linewidth]{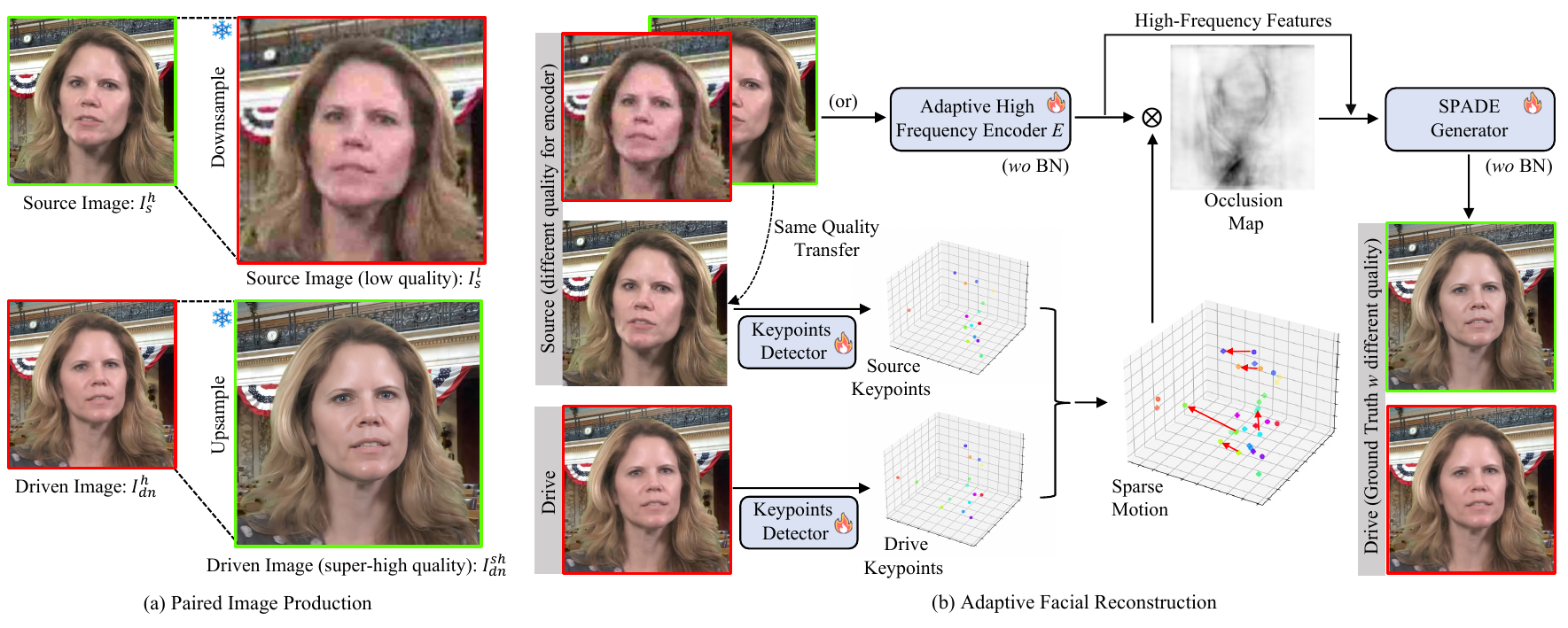}
    \vspace{-0.3cm}
    \caption{
        Illustration of our whole pipeline. (a) We apply pretrained and frozen (snowflake) modules to obtain images of different quality. (2) The pipeline of our framework, the burning represent participation in learnable training. It is worth noting that images with borders of different colors ({\color{green}green} and {\color{red}red}) form a set of training pairs. 
    } \label{fig:framework}
    \vspace{-0.2cm}
\end{figure*}

In recent years, the field of talking head synthesis has seen significant advancements. The primary objective of these systems is to animate a target face using an arbitrary portrait video, whether from the same or a different identity. Some graphics-based neural rendering methods~\cite{koujan2020head2head, song2022everybody,song2021tacr,song2021talking} necessitate both facial geometry and a substantial dataset of target face videos for training. Furthermore, these rendering modules require retraining for different identities or backgrounds. Such methods are limited in their generalizability due to high resource consumption and dependence on 3D geometric data.  In contrast, pure neural rendering methods~\cite{siarohin2019first, zhao2022thin, wang2021one, song2023emotional,song2021fsft} eliminate the need for geometric priors and specialized training videos. Pretrained on large-scale datasets~\cite{nagrani2017voxceleb, chung2018voxceleb2}, these methods transform pixels based on the flow-guided warp or Jacobin matrix~\cite{siarohin2019first} to obtain the driven image. These attributes benefit neural rendering to be widely available and capable of infer the driven videos without finetuning.

One-shot pure neural rendering methods have a broad range of applications, from virtual video conferencing that reduces bandwidth usage~\cite{wang2021one} to character animation in video games and digital human synthesis, \textit{e.t.c.}. As aforementioned, the low quality of the synthesized videos constrains their widespread adoption. The state-of-the-art methods MetaPortrait~\cite{zhang2023metaportrait}, SadTalker~\cite{zhang2023sadtalker} and VideoReTalking~\cite{cheng2022videoretalking} attempt to ameliorate this by retraining an independent super-resolution module to enhance video quality. However, this two-stage synthesis process incurs unnecessary computational overhead and the error accumulation. 

In this work, we propose an adaptive super-resolution method in the talking head generation framework. Inspired by the super-resolution methods, such as ESRGAN~\cite{wang2018esrgan} and Real-ESRGAN~\cite{wang2021realesrgan}, which construct low-quality images data for pairwise training by compressing and downsampling high-quality images. It adaptively captures high-frequency information from low-quality images for reconstruction through a uniquely designed encoder-decoder structure. Additionally, this technique proves compatible with one-shot talking head generation systems. Specifically, we follow the process of pure neural rendering~\cite{wang2021one}. For both the one-shot source face image and the driven images with varying poses, keypoints are extracted separately. Guided by these two sets of keypoints, our encoder-decoder modules extract features from the source image to synthesize novel facial images with varied poses in a fully supervised manner. During training, we intentionally lower the quality of the source image but use high-quality ground-truth images as the supervisory signal. This strategy forces the encoder module to learn the extraction of high-frequency features from low-quality inputs. When we proceed to the inference phase, we employ a high-quality source image. Benifits from the pre-trained enoder's capability to capture these high-frequency features, the system is able to reconstruct images with enhanced clarity.

We validate the efficacy of our approach using both quantitative and qualitative experiments, benchmarking it against existing methods for one-shot talking-head generation. In summary, our key contributions are as follows: 
\begin{itemize}
    \item We introduce a simple but effective method for adaptive high-quality talking head video generation. To our best knowledge, this work is the first to integrate super-resolution techniques into the talking head generation process in an end-to-end fashion. 
    \item Our approach outperforms multiple state-of-the-art methods in both quantitative and qualitative evaluations, as evidenced by tests on large-scale datasets relevant to this task.
\end{itemize}
We expect that our work can provide insights into the future development of high quality portrait video generation and other related tasks.

\begin{figure*}[t]
    \centering
    \includegraphics[width=1.\linewidth]{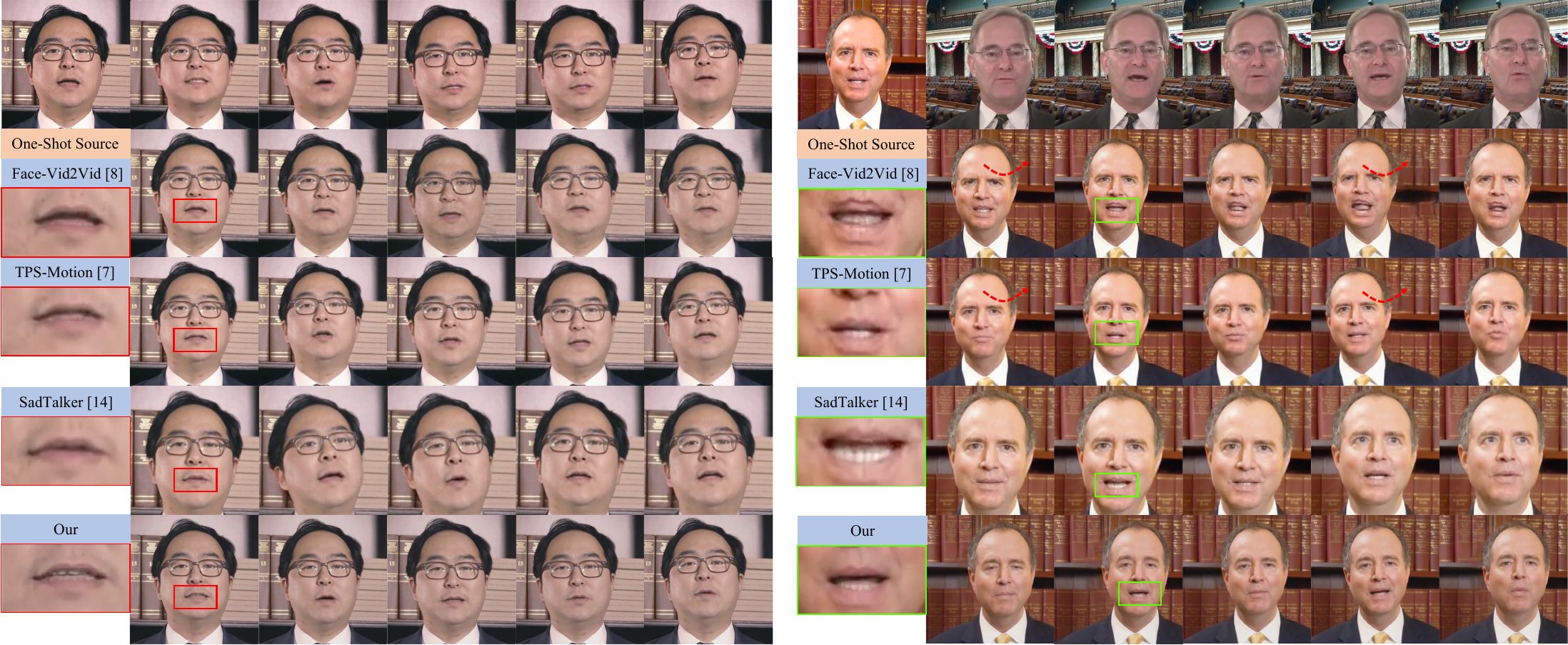}
    \vspace{-0.6cm}
    \caption{
        Qualitative comparison with the baseline methods on the videos from HDTF dataset~\cite{zhang2021flow}. The left part is under the same identity, while the right is cross identity. We zoom in the facial details on the each left. A red arrow indicates incorrect head posture, and the ground truth is on the top. We highly recommend watching our \href{https://www.youtube.com/watch?v=B_-3F51QmKE&t=1s}{supplementary video} for more comparisons.
    } \label{fig:compare}
    \vspace{-0.5cm}
\end{figure*}

\vspace{-0.2cm}
\section{Overview}
\vspace{-0.3cm}

\vspace{-0.1cm}
\subsection{Preliminaries}
\vspace{-0.1cm}
\noindent \textbf{Super-Resolution Network Processing.} The super-resolution networks aim to reconstruct a high-quality image from a low-quality image. In our task, we need to reconstruct high-resolution facial video from low-quality source images. For the given source image $I_{s}^{h}$, the different types of noise, resize methods (area, bilinear, bicubic), and different compression methods for low-quality image creation. Through that, we can obtain the low-quality image $I_{s}^{l}$ paired with the high-quality one $I_{s}^{h}$. 

\noindent \textbf{One-Shot Talking-Head Generation.} The one-shot talking-head generation intent to animate a single source image $I_{s}$ under a series of motion images $\{I_{d1}, I_{d1}, ..., I_{dn}\}$. We can apply the following formula: 
\begin{equation}
\label{equ_0}
    I'_{sn} =  \mathcal{T}(E(I_{s}), I_{dn}),
\end{equation}
the $I'_{sn}$ is the generated portrait image with novel motion but keeps the same identity with $I_{s}$. The $\mathcal{T}$ is the mapping function, and $E$ is the appearance encoder. In this work, we expect E to be able to adaptively capture the high-frequency information from $I_{s}^{l}$, thereby reconstructing a clearer image. It is worth noting that we have ground truth $\{I_{s1}, I_{s1}, ..., I_{sn}\}$ corresponding to $\{I'_{s1}, I'_{s1}, ..., I'_{sn}\}$ for training.

\vspace{-0.3cm}
\subsection{Adaptive High-Frequency Encoder}
\vspace{-0.1cm}
The adaptive high-frequency encoder \( E \) is designed to capture high-frequency details from in-the-wild input sources. Unlike previous methods~\cite{wang2021one, siarohin2019first,zhao2022thin} that focus mainly on pixel transformation, our approach emphasizes the extraction of high-frequency features. To this end, we intentionally downsample the source image \( I_{s1}^{l} \) for training, while maintaining high-quality ground truth images \( I_{sn}^{h} \) for guidance. The process of generating a high-quality image \( I_{sn}^{h} \) from a low-quality image \( I_{s1}^{l} \) can be mathematically described by the formula \( \mathcal{T}(E(I_{s1}^{l}), I_{dn}) \rightarrow I_{sn}^{h} \). In this formulation, \( E \) serves the dual purpose of learning features essential for both facial transformation and texture details.

Moreover, we acknowledge the challenge of differing data distributions between the training and inference phases. To mitigate this, we remove the batch normalization (BN) layers in \( E \), which effectively suppresses artifacts and blurring. Additionally, we employ a pre-trained facial video super-resolution module~\cite{zhang2023metaportrait} to further refine \( I_{sn}^{h} \) to super-high quality \( I_{sn}^{sh} \), thereby forming the training pair \( (I_{s1}^{h}, I_{sn}^{sh}) \). This approach ensures that \( E \) can robustly obtain high-frequency information, not just from low-quality images but also when high-quality images are fed into the system during inference. The pipeline of our framework and cross-quailty training are shown in Fig.~\ref{fig:framework}

\vspace{-0.4cm}
\subsection{Motion Estimation}
\vspace{-0.1cm}
The motion estimation module computes a dense motion field, represented as \(\mathbf{D} \in \mathbb{R}^{3 \times H \times W}\), which aligns a specific frame from the driving video ($\mathbf{D} \in \{I_{d1}, I_{d2}, ..., I_{dn}\}$), with the source frame $I_s$. This motion field is crucial for aligning the feature maps extracted from $I_s$ to the facial pose in $\mathbf{D}$. We define a mapping function $\mathcal{T}_{I_s \leftarrow \mathbf{D}}$ ($\mathbb{R}^{2} \rightarrow \mathbb{R}^{2}$) that correlates each pixel in $\mathbf{D}$ to its corresponding position in $I_s$.

To estimate this dense motion field, we employ a standard U-Net architecture to independently estimate keypoints from both the source image and the driving video. These keypoints are then utilized to calculate sparse motion vectors via a Jacobian matrix through deformation~\cite{siarohin2019first}. A Convolutional Neural Network (CNN), denoted as \(P\), approximates the dense motion field, \(\hat{\mathcal{T}}_{I_s \leftarrow \mathbf{D}}\), serving as a refined transformation function of \(\mathcal{T}_{I_s \leftarrow \mathbf{D}}\) at the keypoint locations. For each keypoint \(p_k\), the heatmap \(\mathbf{H}_k\) is generated, which localizes the regions where the transformations are most relevant. These heatmaps, when concatenated with the sparse motion vectors, assist in predicting occlusion masks. These masks specify which regions of \(\mathbf{D}\) can be generated through warping from \(I_s\), and which should be inpainted. The occlusion masks are applied to images with high-frequency information encoded via the encoder. Finally, a SPADE generator~\cite{park2019SPADE} (without BN layers) is utilized to generate the target video, thereby achieving the desired motion for the source image. 


\vspace{-0.2cm}
\subsection{Training losses}
\vspace{-0.1cm}
Our framework employs multiple loss functions, primarily consisting of two major categories. The first category focuses on facial reconstruction, guiding the facial movement in sparse motion fields; the second category pertains to image quality. 

\noindent\textbf{Facial Structure Losses.} The facial structure losses aim to ensure that both the expressions and poses in the synthesized facial images closely match those in the ground-truth images. These losses comprise various components, including keypoints loss $\mathcal{L}_{k}$, head pose loss $\mathcal{L}_H$, and facial expression loss $\mathcal{L}_E$. The keypoints loss is the L2 distance performed on the 2D keyponits with corresponding depth information (spatial dimension), while both the head pose loss and facial expression loss utilize the L1 distance over the estimated facial parameters.

\noindent\textbf{Equivariance Loss and Deformation Loss.} Following the work of Wang \textit{et al.}~\cite{wang2021one}, we incorporate equivariance loss $\mathcal{L}_{e}$ to ensure the consistency of image-specific keypoints in 2D to 3D transformation. Similarly, deformation loss $\mathcal{L}_{\delta}$ is adopted to constrain canonical space to camera space transformation. 

\noindent\textbf{Perceptual Loss.} Different from the perceptual loss in previous talking-head generation task, the perceptual loss $\mathcal{L}_{P}$ at here is calculated from the feature maps front the activation layers via pretrained VGG network~\cite{simonyan2014very}, which proved in \cite{wang2021realesrgan} to be able to achieve sharper features and consistent lighting. 

\noindent\textbf{GAN Loss.} The multi-resolution patchs GAN loss function $\mathcal{L}_{G}$~\cite{wang2019few, wang2018video}, where the discriminator predicts at the multiple patch-level. And the feature matching loss is also used to optimize the discriminators.

\vspace{-0.4cm}
\section{Experiments}
\vspace{-0.2cm}
\subsection{Experimental Setup}
\vspace{-0.2cm}
\noindent \textbf{Datasets.} We train our model on the large-scale datasets TalkingHead-1KH~\cite{wang2021one} and CelebV-HQ~\cite{zhu2022celebv} respectively, which are two high-quality datasets with face in the center of camera and resolution to $512^2$. Then we eval the performance of the model on the HDTF dataset~\cite{zhang2021flow} and follow the random sampling strategy on the HDTF for evaluation. 

\noindent \textbf{Implementation Details.} We implement our model with PyTorch and four A100 GPUs. To speed up our convergence, the  pretrained model on VoxCeleb-1\cite{chung2018voxceleb2}. The degradation follow the two-stage downsampling in Real-ESRGAN~\cite{wang2021realesrgan}, the image is first down-sampled to $256^2$ and then upsampled to $512^2$. The production of super-resolution images from the pretrained temporal super-resolution~\cite{zhang2023metaportrait} via StyleGAN~\cite{karras2019style}. The keypoints are estimated by Hopenet structure~\cite{Ruiz_2018_CVPR_Workshops}.

\noindent\textbf{Baseline Methods.} For the fairness, we compare our method with the state-of-the-art methods, which include Face-Vid2Vid \cite{wang2021one}, TPS-motion~\cite{zhao2022thin} and SadTalker~\cite{zhang2023sadtalker}. For those methods that use an independently trained super-resolution module for post-processing, we only keep the previous portrait synthesis. We regret not being able to include MetaPortrait~\cite{zhang2023metaportrait}, because its custom video processing part is not public.

\begin{table}[t]
\small
\vspace{-0.25cm}
\begin{center}
\setlength{\tabcolsep}{0.2mm}{
\begin{tabular}{ccccccc}
\hlinew{1.15pt}
\multirow{3}{*}{Methods} &AKD$\downarrow$ &PSNR$\uparrow$ &SSIM$\uparrow$ &FID$\downarrow$ & AED$\downarrow$ &USER$\uparrow$\\
~&&($\times 10^{-2}$) &($\rightarrow 1$) & ($\times 10^{-2}$ ) & &($\rightarrow 5$) \\
\cline{2-7}  
&\multicolumn{6}{c}{TalkingHead-1KH Dataset~\cite{wang2021one}}\\

\hline

Face-Vid2Vid~\cite{wang2021one} & 2.070 & 0.244  & 0.710 & 0.691 & 0.109 & 2.1  \\
TPS-Motion~\cite{zhao2022thin} & 5.041 &0.187  & 0.645 & 0.799 & 0.130 & 1.7   \\
SadTalker~\cite{zhang2023sadtalker} & -- & 0.181 & 0.684 & 0.775 & 0.145 & 1.5  \\
\rowcolor{mygray} Our &\textbf{1.116} &\textbf{0.270} &\textbf{0.783}  & \textbf{0.652}  & \textbf{0.098} & \textbf{3.2} \\
\hline
&\multicolumn{6}{c}{CelebV-HQ Dataset~\cite{zhu2022celebv}}\\
\cline{2-7}
Face-Vid2Vid~\cite{wang2021one} & 2.197 & 0.271  & 0.725 & 0.617 & \textbf{0.097} & 2.0 \\
TPS-Motion~\cite{zhao2022thin} & 5.277 &0.182 & 0.670 & 0.662 & 0.114 & 1.7 \\
SadTalker~\cite{zhang2023sadtalker} & -- & 0.193 & 0.710 & 0.728 & 0.137  & 1.5 \\
\rowcolor{mygray} Our &\textbf{1.308} &\textbf{0.279}  & \textbf{0.790} & \textbf{0.614}  & 0.102 & \textbf{3.1} \\
\hlinew{1.15pt}
\end{tabular}}
\end{center}
\vspace{-0.4cm}
\caption{Quantitative comparison with the baseline methods. The $\downarrow$ ($\uparrow$) indicate that lower (higher) values are indicative of better performance, respectively. The best result are bold.}
\vspace{-0.3cm}
\label{table_all_ab}
\end{table}

\vspace{-0.2cm}
\subsection{Quantitative Evaluation}
\vspace{-0.1cm}

\noindent\textbf{Quantitative Metrics.} Here we mainly focus on three parts, (1) the accuracy of facial motion, (2) the quality of the synthesized images and (3) identity perserving. For the (1), we adopt the average keypoint distance (AKD)~\cite{siarohin2019first}. We measure the PSNR, SSIM and FID for the image quality in (2). The average euclidean distance (AED)~\cite{siarohin2019first} is for the identity of (3). 

\noindent \textbf{Evaluation Results.} The evaluation results are shown in Table~\ref{table_all_ab}. Our method achieves the best performance on the metrics of variety terms, especially on the image quality assessment metrics (PSNR, SSIM and FID), which is due to the adaptive encoder to extract more detailed textures. For the AED metric, our method is able to maintain the best identity information as Face-Vid2Vid~\cite{wang2021one}, although slightly lower in TalkingHead-1KH dataset pretrained. Furthermore, we find that our method was able to achieve the most accurate facial motion in AKD ($1.116 \& 1.308$).

\vspace{-0.2cm}
\subsection{Qualitative Evaluation} 
\vspace{-0.1cm}
\noindent\textbf{Visualization.} We conduct qualitative evaluations to highlight the distinctive advantages of our method over the baseline approaches. We perform our experiments on the same set of facial/head poses or corresponding synchronized audio as the template input with randomly selected source images, which are not involved in training. The qualitative results are visualized in Fig.~\ref{fig:compare}, and the detail of the inner mouth is zoomed in to the right. The portraits generated using our approach exhibit superior teeth quality. In addition, our method has the most accurate head pose than baseline methods (shown in the red arrow in the Fig.~\ref{fig:compare}). We also present the \href{https://www.youtube.com/watch?v=B_-3F51QmKE&t=1s}{supplementary video}\footnote{Youtube: \href{https://www.youtube.com/watch?v=B_-3F51QmKE&t=1s}{https://www.youtube.com/watch?v=B\_-3F51QmKE\&t=1s}}\footnote{BiliBili:  \href{https://www.bilibili.com/video/BV1314y1r7XB/}{https://www.bilibili.com/video/BV1314y1r7XB/}} for dynamic comparison. 

\noindent\textbf{User Study.} We conduct a user study on generated videos, involving 10 participants and a set of 20 videos. In that, we present one video clip and asked participants to respond to the statement ``The video looks real to me" at a time, on a 5-point Likert scale (\textit{1-5, from strongly disagree to strongly agree}). The scores as USER in Table~\ref{table_all_ab}. The user study reveals that our method receives the highest ratings, affirming its effectiveness in human evaluations.

\begin{figure}[t]
    \centering
    \includegraphics[width=1\linewidth]{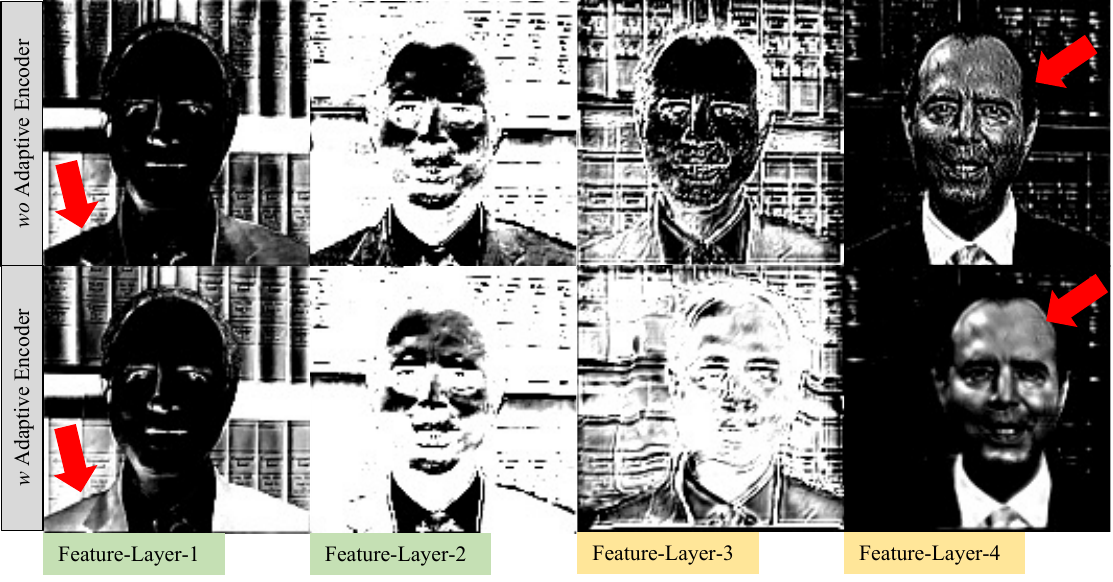}
    \vspace{-0.8cm}
    \caption{
        The visualization of features from each layers in generator \textit{w/wo} adaptive high-frequency encoder $E$. The Feature-Layer-1 and 2 are the features before deformation, Feature-Layer-3 and 4 are the features after deformation.
    } \label{fig:ab}
    \vspace{-0.3cm}
\end{figure}

\vspace{-0.4cm}
\subsection{Ablation Study} 
\vspace{-0.1cm}

\noindent\textbf{Features Visualization.} We visualize the features of the \textit{w/wo} adaptive high-frequency encoder $E$ and the results are shown in Fig.~\ref{fig:ab}. For \textit{wo} $E$, we apply the encoder from the Face-Vid2Vid~\cite{wang2021one} and do not use cross-resolution training. We find that with $E$, the captured features contain more high-frequency information (shown in red arrow in Fig.~\ref{fig:ab}), whether it is before or after being deformed. It is able to incorporate more texture features than low-frequency noise, please zoom in for more details.

\noindent\textbf{Quantitative Evaluation.} The quantitative results can refer to Table~\ref{table_all_ab} (with Face-Vid2Vid~\cite{wang2021one}), it can be found that after the introduction of high-frequency encoder, the obtained image quality is greatly improved on several metrics (PSNR, SSIM and FID).

\vspace{-0.4cm}
\section{Conclusion}
\label{Conclusion}
\vspace{-0.2cm}
In this work, we introduce an adaptive super-resolution approach within the one-shot talking-head video generation domain. Leveraging a simple but effective designed structure, our method is capable of capturing high-frequency details from low-quality images. This allows for the synthesis of high-quality videos without resorting to additional pre-trained modules or a postprocessing. Extensive quantitative and qualitative evaluations on large-scale datasets confirm that our methodology surpasses existing state-of-the-art techniques in high-quality drivable face video generation. 

\vspace{-0.2cm}
\section{Acknowledgments}
\label{Acknowledgments}
\vspace{-0.2cm}
This work was partially supported by the National Science Foundation (NSF) under Grant 1909912 and by the Defense Advance Research Projects Agency (DARPA) under HR00112220003. This paper does not necessarily reflect the position of the Government, and no official endorsement should be inferred.

\small{
\bibliographystyle{IEEEbib}
\bibliography{refs.bib}

\begin{thebibliography}{10}

\bibitem{koujan2020head2head}
Mohammad~Rami Koujan, Michail~Christos Doukas, Anastasios Roussos, and Stefanos Zafeiriou,
\newblock ``Head2head: Video-based neural head synthesis,''
\newblock in {\em 2020 15th IEEE International Conference on Automatic Face and Gesture Recognition (FG 2020)}. IEEE, 2020, pp. 16--23.

\bibitem{song2022everybody}
Linsen Song, Wayne Wu, Chen Qian, Ran He, and Chen~Change Loy,
\newblock ``Everybody’s talkin’: Let me talk as you want,''
\newblock {\em IEEE Transactions on Information Forensics and Security}, vol. 17, pp. 585--598, 2022.

\bibitem{song2021tacr}
Luchuan Song, Bin Liu, Guojun Yin, Xiaoyi Dong, Yufei Zhang, and Jia-Xuan Bai,
\newblock ``Tacr-net: editing on deep video and voice portraits,''
\newblock in {\em Proceedings of the 29th ACM International Conference on Multimedia}, 2021, pp. 478--486.

\bibitem{song2021talking}
Luchuan Song, Bin Liu, and Nenghai Yu,
\newblock ``Talking face video generation with editable expression,''
\newblock in {\em Image and Graphics: 11th International Conference, ICIG 2021, Haikou, China, August 6--8, 2021, Proceedings, Part III 11}. Springer, 2021, pp. 753--764.

\bibitem{siarohin2019first}
Aliaksandr Siarohin, St{\'e}phane Lathuili{\`e}re, Sergey Tulyakov, Elisa Ricci, and Nicu Sebe,
\newblock ``First order motion model for image animation,''
\newblock {\em Advances in neural information processing systems}, vol. 32, 2019.

\bibitem{zhao2022thin}
Jian Zhao and Hui Zhang,
\newblock ``Thin-plate spline motion model for image animation,''
\newblock in {\em Proceedings of the IEEE/CVF Conference on Computer Vision and Pattern Recognition}, 2022, pp. 3657--3666.

\bibitem{wang2021one}
Ting-Chun Wang, Arun Mallya, and Ming-Yu Liu,
\newblock ``One-shot free-view neural talking-head synthesis for video conferencing,''
\newblock in {\em Proceedings of the IEEE/CVF conference on computer vision and pattern recognition}, 2021, pp. 10039--10049.

\bibitem{song2023emotional}
Luchuan Song, Guojun Yin, Zhenchao Jin, Xiaoyi Dong, and Chenliang Xu,
\newblock ``Emotional listener portrait: Neural listener head generation with emotion,''
\newblock in {\em Proceedings of the IEEE/CVF International Conference on Computer Vision}, 2023, pp. 20839--20849.

\bibitem{song2021fsft}
Luchuan Song, Guojun Yin, Bin Liu, Yuhui Zhang, and Nenghai Yu,
\newblock ``Fsft-net: face transfer video generation with few-shot views,''
\newblock in {\em 2021 IEEE International Conference on Image Processing (ICIP)}. IEEE, 2021, pp. 3582--3586.

\bibitem{nagrani2017voxceleb}
Arsha Nagrani, Joon~Son Chung, and Andrew Zisserman,
\newblock ``Voxceleb: a large-scale speaker identification dataset,''
\newblock {\em arXiv preprint arXiv:1706.08612}, 2017.

\bibitem{chung2018voxceleb2}
Joon~Son Chung, Arsha Nagrani, and Andrew Zisserman,
\newblock ``Voxceleb2: Deep speaker recognition,''
\newblock {\em arXiv preprint arXiv:1806.05622}, 2018.

\bibitem{zhang2023metaportrait}
Bowen Zhang, Chenyang Qi, Pan Zhang, Bo~Zhang, HsiangTao Wu, Dong Chen, Qifeng Chen, Yong Wang, and Fang Wen,
\newblock ``Metaportrait: Identity-preserving talking head generation with fast personalized adaptation,''
\newblock in {\em Proceedings of the IEEE/CVF Conference on Computer Vision and Pattern Recognition}, 2023, pp. 22096--22105.

\bibitem{zhang2023sadtalker}
Wenxuan Zhang, Xiaodong Cun, Xuan Wang, Yong Zhang, Xi~Shen, Yu~Guo, Ying Shan, and Fei Wang,
\newblock ``Sadtalker: Learning realistic 3d motion coefficients for stylized audio-driven single image talking face animation,''
\newblock in {\em Proceedings of the IEEE/CVF Conference on Computer Vision and Pattern Recognition}, 2023, pp. 8652--8661.

\bibitem{cheng2022videoretalking}
Kun Cheng, Xiaodong Cun, Yong Zhang, Menghan Xia, Fei Yin, Mingrui Zhu, Xuan Wang, Jue Wang, and Nannan Wang,
\newblock ``Videoretalking: Audio-based lip synchronization for talking head video editing in the wild,''
\newblock in {\em SIGGRAPH Asia 2022 Conference Papers}, 2022, pp. 1--9.

\bibitem{wang2018esrgan}
Xintao Wang, Ke~Yu, Shixiang Wu, Jinjin Gu, Yihao Liu, Chao Dong, Yu~Qiao, and Chen Change~Loy,
\newblock ``Esrgan: Enhanced super-resolution generative adversarial networks,''
\newblock in {\em Proceedings of the European conference on computer vision (ECCV) workshops}, 2018, pp. 0--0.

\bibitem{wang2021realesrgan}
Xintao Wang, Liangbin Xie, Chao Dong, and Ying Shan,
\newblock ``Real-esrgan: Training real-world blind super-resolution with pure synthetic data,''
\newblock in {\em International Conference on Computer Vision Workshops (ICCVW)}.

\bibitem{zhang2021flow}
Zhimeng Zhang, Lincheng Li, Yu~Ding, and Changjie Fan,
\newblock ``Flow-guided one-shot talking face generation with a high-resolution audio-visual dataset,''
\newblock in {\em Proceedings of the IEEE/CVF Conference on Computer Vision and Pattern Recognition}, 2021, pp. 3661--3670.

\bibitem{park2019SPADE}
Taesung Park, Ming-Yu Liu, Ting-Chun Wang, and Jun-Yan Zhu,
\newblock ``Semantic image synthesis with spatially-adaptive normalization,''
\newblock in {\em Proceedings of the IEEE Conference on Computer Vision and Pattern Recognition}, 2019.

\bibitem{simonyan2014very}
Karen Simonyan and Andrew Zisserman,
\newblock ``Very deep convolutional networks for large-scale image recognition,''
\newblock {\em arXiv preprint arXiv:1409.1556}, 2014.

\bibitem{wang2019few}
Ting-Chun Wang, Ming-Yu Liu, Andrew Tao, Guilin Liu, Jan Kautz, and Bryan Catanzaro,
\newblock ``Few-shot video-to-video synthesis,''
\newblock {\em arXiv preprint arXiv:1910.12713}, 2019.

\bibitem{wang2018video}
Ting-Chun Wang, Ming-Yu Liu, Jun-Yan Zhu, Guilin Liu, Andrew Tao, Jan Kautz, and Bryan Catanzaro,
\newblock ``Video-to-video synthesis,''
\newblock {\em arXiv preprint arXiv:1808.06601}, 2018.

\bibitem{zhu2022celebv}
Hao Zhu, Wayne Wu, Wentao Zhu, Liming Jiang, Siwei Tang, Li~Zhang, Ziwei Liu, and Chen~Change Loy,
\newblock ``Celebv-hq: A large-scale video facial attributes dataset,''
\newblock in {\em European conference on computer vision}. Springer, 2022, pp. 650--667.

\bibitem{karras2019style}
Tero Karras, Samuli Laine, and Timo Aila,
\newblock ``A style-based generator architecture for generative adversarial networks,''
\newblock in {\em Proceedings of the IEEE/CVF conference on computer vision and pattern recognition}, 2019, pp. 4401--4410.

\bibitem{Ruiz_2018_CVPR_Workshops}
Nataniel Ruiz, Eunji Chong, and James~M. Rehg,
\newblock ``Fine-grained head pose estimation without keypoints,''
\newblock in {\em The IEEE Conference on Computer Vision and Pattern Recognition (CVPR) Workshops}, June 2018.

\end{thebibliography}
}

\end{document}